\documentclass[letterpaper,10pt,conference]{ieeeconf}

\IEEEoverridecommandlockouts
\usepackage{cite}
\usepackage{amsmath,amssymb,amsfonts}
\usepackage{graphicx}
\usepackage{xcolor}
\usepackage[T1]{fontenc}
\usepackage{booktabs}
\usepackage{url}
\usepackage{tikz}
\usetikzlibrary{arrows.meta,positioning}

\begin{document}

\title{
Can Synthetic UAV Benchmarks Be Trusted?\\
Real--Synthetic Validation for UAV Model Selection
}

\author{
Yuyi Zhi$^{1,\dagger}$, Min Du$^{2,\dagger}$,
Bill Shi$^{2,\dagger}$, Guodong Yuan$^{2}$\\
Junheng Liang$^{3}$, Peifan Liao$^{4}$
and Xin Guan$^{2,*}$%
\thanks{This work has been submitted to IEEE for possible
publication. Copyright may be transferred without notice,
after which this version may no longer be accessible.}%
\thanks{$^{\dagger}$These authors contributed equally to this work.}%
\thanks{$^{1}$Yuyi Zhi is with Zhejiang University.
Email: \texttt{zhiyuyi@zju.edu.cn}.}%
\thanks{$^{2}$Min Du, Bill Shi, Guodong Yuan,
and Xin Guan are with Avant Robotics.
Emails: \texttt{dumin2712220318@gmail.com},
\texttt{tianyu.s@outlook.com},
\texttt{yuanguodong@autel.com},
and \texttt{garguan2001@outlook.com}, respectively.}%
\thanks{$^{3}$Junheng Liang is with
University of Science and Technology Beijing.
Email: \texttt{vincent\_leon@xs.ustb.edu.cn}.}%
\thanks{$^{4}$Peifan Liao is with Tsinghua University.
Email: \texttt{liaopf24@mails.tsinghua.edu.cn}.}%
\thanks{$^{*}$Corresponding author: Xin Guan
(\texttt{garguan2001@outlook.com}).}%
}

\maketitle

\begin{abstract}
Synthetic UAV imagery can expand pre-deployment test evidence when
task-specific Real data are scarce, but visual plausibility or semantic label
alignment alone does not establish validity for model selection. We study when
Synthetic evaluation preserves the model-selection conclusions obtained from
Real UAV data through two complementary cases under fixed evaluation
protocols. In a requirement-specific road-debris benchmark, four
language-conditioned perception systems obtain lower absolute F1 on Synthetic
data, yet the point-estimate ordering and top-ranked model are preserved;
bootstrap analysis yields no supported pairwise reversal, and size-matched
resampling preserves the top-model decision. In contrast, a closed-set stress test on a separately materialized business
corpus, using four VisDrone-compatible detectors over eight empirically
supported classes, changes the point-estimate top-ranked detector and reverses three of
six pairwise point-estimate preference directions. A complementary audit of five UAV
business scenarios further reveals substantial Real--Synthetic
mismatch in class composition, object density, and target scale,
including systematic under-representation of small targets. These results
show that Synthetic UAV benchmark trust is conditional: semantic coverage is
insufficient, and validity must be established empirically for the task,
operating regime, and model panel of interest.
\end{abstract}


\section{Introduction}
\label{sec:introduction}

Selecting a perception system for a UAV inspection task requires test
evidence that reflects the intended operation. For task-specific events such
as road debris, however, only a small Real test set may be available before
deployment. Public datasets such as VisDrone provide valuable aerial
perception benchmarks~\cite{zhu2022visdrone}, but do not cover every
inspection requirement. Synthetic imagery offers a way to expand this
evidence, provided that it supports the same model-selection decision as the
available Real reference.

Recent aerial-generation methods improve layout control, object
placement, and small-object synthesis~\cite{tang2025aerogen,li2026uavgen}.
However, most synthetic UAV data are evaluated primarily as training or
augmentation resources: their utility is established by whether models trained
with synthetic data perform better on real imagery. Useful training data are
not necessarily valid test data. More broadly, prior work on synthetic model evaluation shows that agreement
with real-world errors or model rankings must be established
empirically~\cite{shoshan2023syntheticselection,
kang2025syntheticbenchmark,breugel2023synthetictest,
boyeau2025autoeval}. A visually plausible synthetic
test set may still shift absolute scores, distort task difficulty, or change
which model would be preferred.

This paper therefore asks: \emph{When can a synthetic UAV benchmark be trusted
for model selection, and when does it fail?} We use ``trust'' in a deliberately
narrow empirical sense: whether evaluation on Synthetic data preserves the
model-selection conclusion obtained from Real UAV data under a fixed,
pre-specified protocol. Trust does not imply numerical score equivalence, operational safety, or
certification. Accordingly, we evaluate model-selection agreement directly
and separately examine whether the constructed images reflect relevant
properties of the Real reference data.

\begin{figure*}[t]
    \centering
    \includegraphics[width=0.98\textwidth]{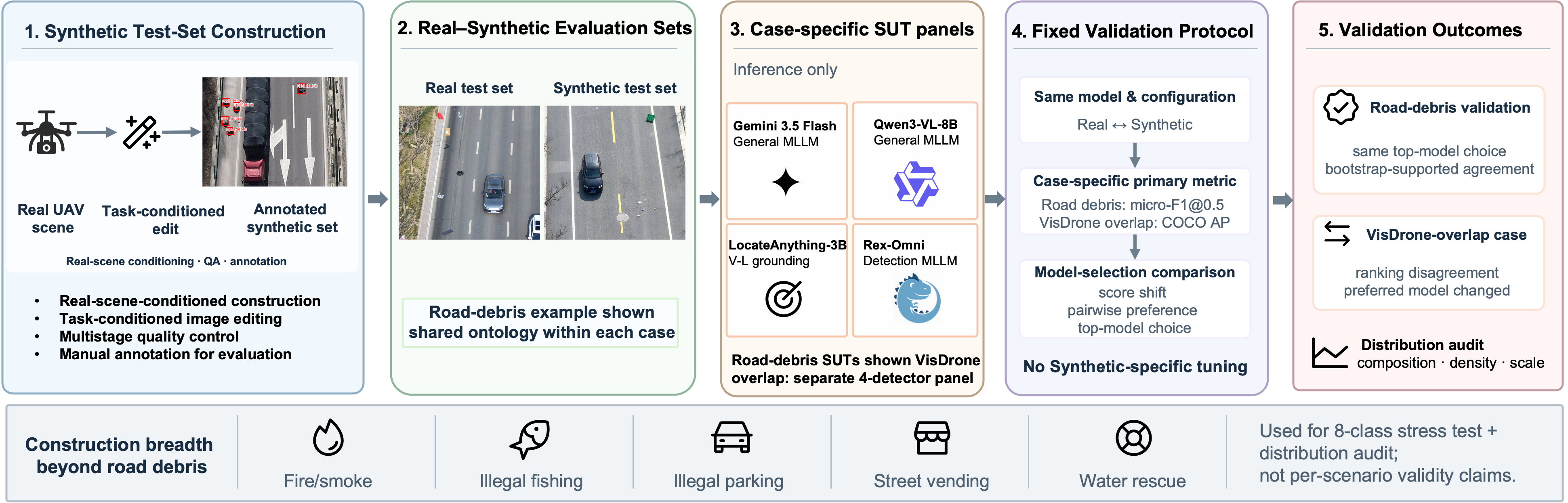}
    \caption{
Overview of the Real--Synthetic validation study. The construction example
and pictured SUTs refer to road debris; the VisDrone-overlap test uses a
separately materialized corpus and a distinct detector panel. The two cases
yield different model-selection outcomes, while the five-scenario audit
documents distributional discrepancies in the business corpus.
}
    \label{fig:overview}
\end{figure*}

Figure~\ref{fig:overview} summarizes the study. In the road-debris case,
all four language-conditioned systems obtain lower F1 on Synthetic data,
yet the top-model choice is unchanged under bootstrap and size-matched
resampling. In the VisDrone-overlap test, all four closed-set detectors
obtain higher Synthetic AP, but the point-estimate preferred detector
changes. We complement these comparisons with a separate audit of class
composition, object density, and target scale in five business-scenario
collections. The cases differ in task, reference data, model panel, and generation backend;
their contrast is empirical rather than a controlled ablation.

Our main contributions are:

\begin{itemize}
\item \textbf{Case-based Real--Synthetic model-selection evaluation.}
We compare two complementary UAV perception cases under fixed evaluation
protocols: a requirement-specific road-debris benchmark that preserves the
top-model choice, and a VisDrone-overlap closed-set test in which the
point-estimate preferred detector changes.

\item \textbf{Auditable requirement-specific test-set construction.}
We construct Synthetic UAV evaluation data from Real scene content using
task-conditioned materialization, multistage quality control, manual
annotation, and recorded provenance.

\item \textbf{Cross-scenario distributional audit.}
We quantify differences in annotated class prevalence, object density, and
target scale across five UAV business scenarios, identifying distributional
discrepancies that shared semantic labels do not capture.
\end{itemize}

\section{Related Work}
\label{sec:related-work}

\noindent\textbf{Synthetic data for aerial and UAV perception.}
Synthetic aerial datasets have been developed through both simulation and
generative pipelines. Mid-Air, SynDrone, and SkyScenes provide controllable
UAV or aerial imagery for perception and synthetic-to-real studies
\cite{fonder2019midair,rizzoli2023syndrone,khose2024skyscenes}, while recent
generative methods such as AeroGen and UAVGen improve control over object
placement, appearance, and small-object synthesis
\cite{tang2025aerogen,li2026uavgen}. These works primarily establish synthetic
data as a resource for training, augmentation, or domain transfer. Our focus
is different: we treat synthetic imagery as a \emph{test set} and ask whether
it preserves the model-selection conclusions obtained from Real UAV data.

\noindent\textbf{Synthetic evaluation and model-selection validity.}
Synthetic imagery has also been used for controlled evaluation and model
assessment. Prior UAV and vision benchmarks use synthetic scenes to probe
detector behavior under controlled conditions
\cite{lee2021uavvalidation,shen2023archangel,foszner2023crowdsim2}, while
broader studies examine synthetic test data for model selection and efficient
model evaluation
\cite{shoshan2023syntheticselection,kang2025syntheticbenchmark,
breugel2023synthetictest,boyeau2025autoeval}. Task Me Anything and scenario-driven generation further
explore controllable construction of evaluation cases
\cite{zhang2024taskmeanything,xu2026scenariodriven}.
These studies motivate evaluating synthetic benchmarks through downstream
measurement agreement rather than visual realism alone. We extend this view by
examining both a case in which the model-selection decision is preserved and a
counterexample in which a shared-ontology Synthetic evaluation changes the
preferred model. For the requirement-specific road-debris task, open-vocabulary
and language-conditioned systems such as OWL-ViT, GLIP, Grounding DINO,
Rex-Omni, and LocateAnything provide suitable localization interfaces
\cite{minderer2022owlvit,li2022glip,liu2024groundingdino,
jiang2026rexomni,wang2026locateanything}.

\begin{figure*}[!t]
    \centering
    \includegraphics[width=0.94\textwidth]{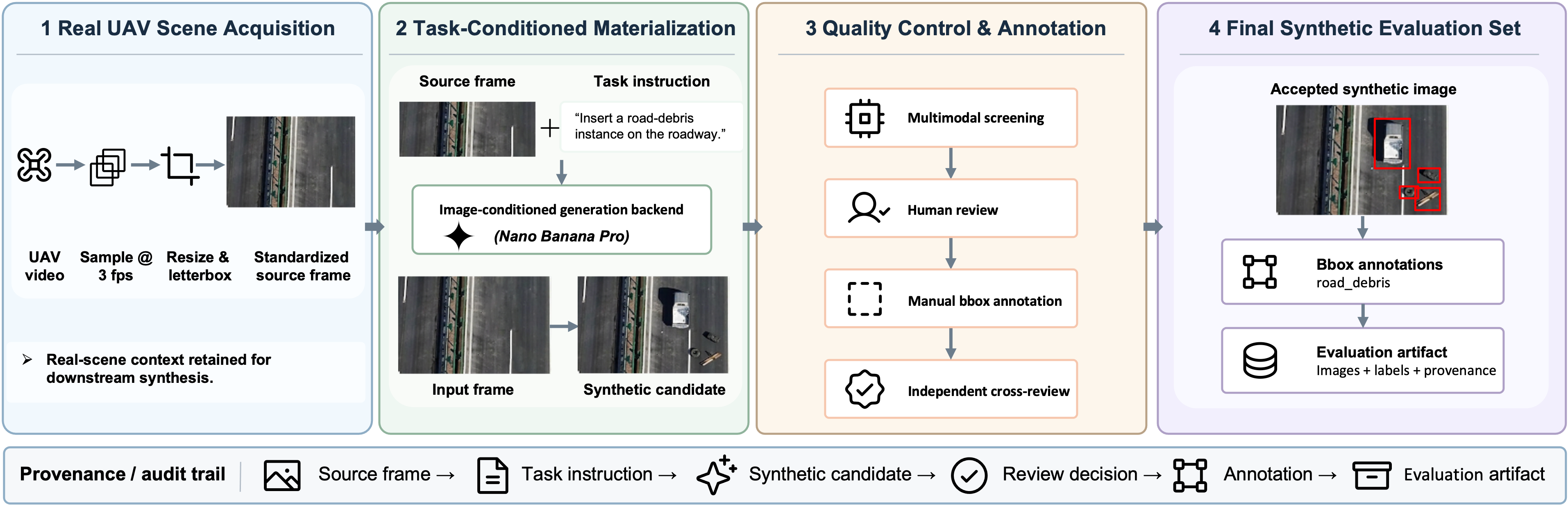}
    \caption{
    Road-debris Synthetic test-set construction pipeline.
    A Real UAV source frame and task instruction condition image generation,
    followed by multimodal screening, human review, manual bounding-box
    annotation, independent cross-review, and provenance recording before
    admission to the final evaluation set.
    }
    \label{fig:pipeline}
\end{figure*}

\noindent\textbf{Real UAV benchmarks and operating-regime grounding.}
Real UAV benchmarks demonstrate that aerial perception varies substantially
with viewpoint, target scale, scene composition, sensing condition, and
operational context. VisDrone and AU-AIR cover drone-view traffic perception
\cite{zhu2022visdrone,bozcan2020auair}, while SeaDronesSee, HIT-UAV,
UAVScenes, and ATR-UMOD extend evaluation to maritime rescue, thermal sensing,
multimodal perception, and diverse operating conditions
\cite{varga2022seadronessee,suo2023hituav,wang2025uavscenes,
chen2025atrumod}. Motivated by this diversity, we use Real UAV data not only
as a performance reference but also to audit whether Synthetic collections
cover task-relevant operating regimes. In particular, we compare class composition, object density,
and target scale across five business-aligned UAV scenario families. These diagnostics document distributional discrepancies rather than
cross-scenario model-selection validity by themselves.

\section{Requirement-Specific Road-Debris Benchmark Construction}
\label{sec:construction}
This section details the construction pipeline used for the road-debris
benchmark, for which the production process and provenance are available to
the authors. The goal is to construct requirement-specific Synthetic test
artifacts while keeping generation, quality control, annotation, and
downstream validation conceptually separate. Construction determines which
artifacts enter the evaluation set; whether they support valid model selection
is assessed independently in Sec.~\ref{sec:experiments} using the protocol of
Sec.~\ref{sec:protocol}.

\subsection{Requirement- and Scene-Conditioned Materialization}
\label{sec:materialization}

For the road-debris benchmark, source frames are sampled at 3~fps from
UAV videos captured with an Autel EVO MAX across multiple urban road
environments. The source videos span flight altitudes of 30--80~m, zoom
factors of $1$--$5\times$, gimbal pitch angles of
$-45^\circ$, $-60^\circ$, and $-90^\circ$, and native resolutions from
1080P to 4K. Each sampled frame is isotropically resized and center-padded
to $1440\times1080$ pixels before generation.

For a standardized source frame $x_i$ and target condition $c$, a
task-specific instruction $p_{i,c}$ specifies the requested target,
appearance, and placement. Candidate Synthetic images are produced with
Nano Banana Pro using image-conditioned generation,
\begin{equation}
\tilde{x}_{i,c}
=
G_{\mathrm{NB}}(x_i,p_{i,c}),
\label{eq:generation}
\end{equation}
where $G_{\mathrm{NB}}$ denotes the deployed generation backend. When a
corresponding normal structure is already present in the source image, the
instruction may request an edit of that structure rather than insertion of a
new object.

Using a Real source frame conditions generation on an observed aerial
viewpoint, background, and scene layout; it does not guarantee that local
geometry or other image statistics remain unchanged. The generator is treated
as a black-box materialization operator and is not trained or adapted using
downstream SUT results. The five-scenario business corpus was materialized separately using a
different internal generation backend; its detailed production pipeline is
not available to the authors.

\subsection{Quality Control, Annotation, and Artifact Packaging}
\label{sec:quality-control}

Generated candidates undergo sequential automatic and human quality control
before annotation. A multimodal screening model first rejects candidates with
severe blur, implausible target structure, obvious generation artifacts,
insufficient realization of the requested condition, or strongly exaggerated
target appearance. Candidates that pass this stage are then reviewed
image-by-image by a human reviewer.

Accepted images are manually annotated with bounding boxes. Each image has
one initial annotator, after which an independent reviewer checks category
correctness, missing instances, box completeness, and localization accuracy.
Annotations that fail cross-review are revised and reviewed again before
admission. Let $q_i^M$, $q_i^H$, and $q_i^C$ denote automatic screening,
human image review, and annotation cross-review, respectively. Final admission
is

\begin{equation}
a_i=q_i^M q_i^H q_i^C,
\qquad
a_i=1
\Rightarrow
(\tilde{x}_i,Y_i,m_i)\in\mathcal D_{\mathrm{eval}},
\label{eq:admission}
\end{equation}

where $Y_i$ denotes the final bounding-box annotations and $m_i$ stores the
available provenance record. Accepted annotations are exported in COCO and
YOLO formats for downstream evaluation.

Figure~\ref{fig:pipeline} summarizes the road-debris construction pipeline,
including source-frame conditioning, task-conditioned materialization,
multistage review, annotation, and artifact packaging.

\section{Real--Synthetic Model-Selection Validation Protocol}
\label{sec:protocol}

We evaluate each Synthetic benchmark against a declared Real reference under
a fixed, pre-specified protocol, separating downstream model-selection
fidelity from dataset-level distributional diagnostics.

\subsection{Fixed Cross-Benchmark Evaluation}
\label{sec:aligned-eval}

For evaluation case $d$, let $\mathcal D_d^{\mathrm R}$ and
$\mathcal D_d^{\mathrm S}$ denote the Real reference and Synthetic test
collections, respectively. Their native annotations are mapped to a
pre-specified shared evaluation ontology $\mathcal Y_d$ before model outcomes
are compared. The road-debris case uses the single canonical
\texttt{road\_debris} category, whereas the closed-set stress test uses the
predefined subset of business-corpus categories that overlap the VisDrone
detection ontology~\cite{zhu2022visdrone}. This shared label space provides a
common scoring interface; it does not imply that the underlying scene
distributions are equivalent.

Each case uses a fixed SUT panel
$\mathcal M_d=\{M_1,\ldots,M_{K_d}\}$. For system $M_k$, its checkpoint or
API identifier, task input, preprocessing, coordinate conversion,
post-processing policy, class mapping, evaluator, and software configuration
$\phi_{k,d}$ are specified before the Real--Synthetic comparison and kept
unchanged across the two test sets. No Synthetic-specific fine-tuning,
checkpoint replacement, prompt selection, test-time adaptation, confidence
calibration, or result-dependent inference modification is permitted.

Let $g_d$ denote the primary scalar model-selection metric for case $d$. We use
micro-F1 at IoU $=0.5$ for road debris and COCO AP over IoU thresholds
$0.50{:}0.05{:}0.95$ for the VisDrone-overlap closed-set evaluation. For road debris, prediction--ground-truth matching is performed independently
within each image. All prediction--GT pairs are sorted by descending IoU, and
a pair is greedily accepted when its IoU is at least $0.5$ and neither box has
been matched previously. The assignment is therefore one-to-one and does not
use confidence scores. Unmatched predictions and ground-truth boxes are
counted as false positives and false negatives, respectively.

Unparseable model responses are retained as empty prediction sets rather than
excluded from evaluation. For partially valid responses, boxes that pass the
deterministic parsing and validity checks are retained. No ground-truth image
is removed because of an output-format failure. Counts are pooled over all
images before computing
\begin{equation}
F_1=\frac{2TP}{2TP+FP+FN}.
\label{eq:micro-f1}
\end{equation}
The corresponding Real and Synthetic scores and their absolute shift are

\begin{equation}
\begin{aligned}
s_{k,d}^{\mathrm R}
&=
g_d(M_k,\mathcal D_d^{\mathrm R};\phi_{k,d}),\\
s_{k,d}^{\mathrm S}
&=
g_d(M_k,\mathcal D_d^{\mathrm S};\phi_{k,d}),\\
\Delta_{k,d}
&=
s_{k,d}^{\mathrm S}-s_{k,d}^{\mathrm R}.
\end{aligned}
\label{eq:score-gap}
\end{equation}

Secondary metrics are retained for diagnosis rather than used to redefine the
model-selection criterion. These include Precision, Recall, matched-box IoU, and output behavior for
road debris, and AP50, AP75, and AR@300 for the closed-set stress test. A small $\Delta_{k,d}$ alone is not treated as
benchmark equivalence.

\subsection{Model-Selection Fidelity and Finite-Sample Uncertainty}
\label{sec:agreement}

For systems $M_i$ and $M_j$, model preference is represented by the
benchmark-specific score margins

\begin{equation}
\delta_{ij,d}^{\mathrm R}
=
s_{i,d}^{\mathrm R}-s_{j,d}^{\mathrm R},
\qquad
\delta_{ij,d}^{\mathrm S}
=
s_{i,d}^{\mathrm S}-s_{j,d}^{\mathrm S}.
\label{eq:pairwise-preference}
\end{equation}

\begin{figure}[!htbp]
\centering

\resizebox{0.98\columnwidth}{!}{%
\begin{tikzpicture}[
    font=\footnotesize,
    axis/.style={black!55, line width=0.5pt},
    ci/.style={black, line width=1.2pt},
    point/.style={circle, fill=black, inner sep=1.5pt},
    decision/.style={align=center}
]

\node[font=\bfseries] at (3.0,5.1) {Real margin};
\node[font=\bfseries] at (6.6,5.1) {Synthetic margin};
\node[font=\bfseries] at (9.7,5.1) {Decision};

\node[font=\bfseries, anchor=west] at (0.0,4.0) {Preserved};

\draw[axis] (1.8,3.8) -- (4.2,3.8);
\draw[axis] (3.0,3.65) -- (3.0,3.95);
\node[below] at (3.0,3.65) {$0$};

\draw[ci] (3.25,3.8) -- (4.00,3.8);
\draw[ci] (3.25,3.68) -- (3.25,3.92);
\draw[ci] (4.00,3.68) -- (4.00,3.92);
\node[point] at (3.63,3.8) {};

\draw[axis] (5.4,3.8) -- (7.8,3.8);
\draw[axis] (6.6,3.65) -- (6.6,3.95);
\node[below] at (6.6,3.65) {$0$};

\draw[ci] (6.85,3.8) -- (7.60,3.8);
\draw[ci] (6.85,3.68) -- (6.85,3.92);
\draw[ci] (7.60,3.68) -- (7.60,3.92);
\node[point] at (7.23,3.8) {};

\node[decision] at (9.7,3.8) {same supported direction};

\node[font=\bfseries, anchor=west] at (0.0,2.5) {Reversed};

\draw[axis] (1.8,2.3) -- (4.2,2.3);
\draw[axis] (3.0,2.15) -- (3.0,2.45);
\node[below] at (3.0,2.15) {$0$};

\draw[ci] (3.25,2.3) -- (4.00,2.3);
\draw[ci] (3.25,2.18) -- (3.25,2.42);
\draw[ci] (4.00,2.18) -- (4.00,2.42);
\node[point] at (3.63,2.3) {};

\draw[axis] (5.4,2.3) -- (7.8,2.3);
\draw[axis] (6.6,2.15) -- (6.6,2.45);
\node[below] at (6.6,2.15) {$0$};

\draw[ci] (5.60,2.3) -- (6.35,2.3);
\draw[ci] (5.60,2.18) -- (5.60,2.42);
\draw[ci] (6.35,2.18) -- (6.35,2.42);
\node[point] at (5.98,2.3) {};

\node[decision] at (9.7,2.3) {opposite supported directions};

\node[font=\bfseries, anchor=west] at (0.0,1.0) {Unresolved};

\draw[axis] (1.8,0.8) -- (4.2,0.8);
\draw[axis] (3.0,0.65) -- (3.0,0.95);
\node[below] at (3.0,0.65) {$0$};

\draw[ci] (2.45,0.8) -- (3.65,0.8);
\draw[ci] (2.45,0.68) -- (2.45,0.92);
\draw[ci] (3.65,0.68) -- (3.65,0.92);
\node[point] at (3.25,0.8) {};

\draw[axis] (5.4,0.8) -- (7.8,0.8);
\draw[axis] (6.6,0.65) -- (6.6,0.95);
\node[below] at (6.6,0.65) {$0$};

\draw[ci] (6.85,0.8) -- (7.60,0.8);
\draw[ci] (6.85,0.68) -- (6.85,0.92);
\draw[ci] (7.60,0.68) -- (7.60,0.92);
\node[point] at (7.23,0.8) {};

\node[decision] at (9.7,0.8) {at least one interval\\includes zero};

\node[font=\scriptsize] at (5.0,-0.05)
{dot = point estimate; horizontal bar = 95\% bootstrap interval};

\end{tikzpicture}%
}

\caption{
Pairwise Real--Synthetic model-selection fidelity.
Black dots denote point-estimate performance margins and horizontal bars
denote 95\% bootstrap intervals. Preserved, reversed, and unresolved outcomes
are determined by the interval positions relative to zero.
}
\label{fig:pairwise-protocol}

\end{figure}
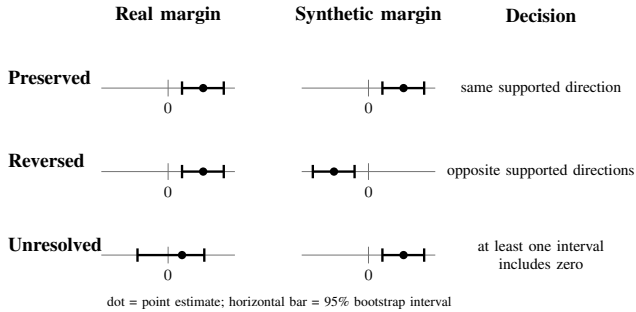

Positive and negative margins indicate preference for $M_i$ and $M_j$,
respectively. Figure~\ref{fig:pairwise-protocol} illustrates the three
uncertainty-aware outcomes. 

A pair is \emph{preserved} when the Real and
Synthetic intervals both exclude zero and support the same preference
direction, \emph{reversed} when they exclude zero in opposite directions, and
\emph{unresolved} when either interval includes zero.

For the road-debris case, finite-sample uncertainty is estimated from cached
predictions using 10,000 image-level nonparametric bootstrap replicates with
replacement. Real and Synthetic collections are resampled independently
because their images are not instance-paired. Within each benchmark, however,
the same resampled image indices are used for every SUT so that pairwise
comparisons preserve their dependence structure. Micro-F1 is recomputed from
pooled TP, FP, and FN counts in every replicate. We report nominal
per-comparison 95\% percentile bootstrap confidence intervals for pairwise
score margins, together with bootstrap top-1 frequency.

When the Real and Synthetic collections differ materially in size, we also
perform a size-matched sensitivity analysis by repeatedly subsampling the
larger collection to the smaller image count without replacement, using the
same subset for every SUT within a repetition. This analysis tests whether a
model-selection conclusion depends primarily on unequal benchmark size and is
kept separate from the bootstrap-based
preserved/reversed/unresolved classification.
In the present study, bootstrap and size-matched resampling are applied only
to the road-debris case. The VisDrone-overlap stress test is reported at the
point-estimate level and is therefore not assigned bootstrap-based
preserved/reversed/unresolved labels.

\subsection{Distributional Audit}
\label{sec:distribution-audit}

Model-selection agreement is complemented by a dataset-level audit of the
broader 45-category business corpus across five UAV scenario families.
For each family, we compare Real and Synthetic class support and composition,
annotated-object density, and target scale. Target scale is represented by normalized bounding-box
area,
$a_{\mathrm{norm}}=(w/W)(h/H)$, so that images with different native
resolutions remain comparable. Each Real family's first-quartile
($Q_{25}$) normalized area is additionally used as a reference threshold for
quantifying Synthetic coverage of the corresponding small-target regime.

These measurements diagnose whether semantic coverage is accompanied by
similar operating-regime support. They are not treated as sufficient evidence
of model-selection validity, nor are observed distributional mismatches
interpreted causally as the source of a particular score gap or preference
reversal.

\section{Experiments}
\label{sec:experiments}

\subsection{Experimental Setup}
\label{sec:experimental-setup}

\paragraph{Road-debris case}
The road-debris experiment uses independently collected Real and Synthetic UAV
test sets under the same canonical \texttt{road\_debris} bounding-box
ontology. The Real set contains 76 images with 417 annotated instances and the
Synthetic set contains 200 images with 1,101 instances. Their mean object
densities are similar (5.49 versus 5.51 instances/image). Neither collection
is used for training, fine-tuning, prompt selection, or test-time adaptation.

We evaluate four pretrained language-conditioned perception systems:
Gemini 3.5 Flash,\footnote{Google DeepMind Gemini 3.5 Flash model card:
\url{https://deepmind.google/models/model-cards/gemini-3-5-flash/}}
Qwen3-VL-8B-Instruct~\cite{bai2025qwen3vl},
LocateAnything-3B~\cite{wang2026locateanything}, and
Rex-Omni~\cite{jiang2026rexomni}. Table~\ref{tab:sut-panel} summarizes their fixed task interfaces.

\begin{table}[!htbp]
\centering
\caption{Road-debris SUT panel and fixed task interfaces.}
\label{tab:sut-panel}
\scriptsize
\setlength{\tabcolsep}{2pt}
\renewcommand{\arraystretch}{1.08}

\begin{tabular}{@{}
p{0.30\columnwidth}
p{0.25\columnwidth}
p{0.34\columnwidth}
@{}}
\toprule
Model & System type & Task interface \\
\midrule

Gemini 3.5 Flash
& General MLLM
& Road-debris instruction \\

Qwen3-VL-8B
& General MLLM
& Road-debris instruction \\

LocateAnything-3B
& V-L grounding
& \texttt{road debris} \\

Rex-Omni
& Detection MLLM
& \texttt{road debris} \\

\bottomrule
\end{tabular}
\end{table}

The panel spans general-purpose multimodal and localization-oriented
vision-language systems and is used for benchmark validation rather than
leaderboard comparison. Each model-specific task input, preprocessing,
output parser, coordinate conversion, and deterministic post-processing
policy is fixed before evaluation and kept unchanged between Real and
Synthetic data.
Micro-averaged F1 at IoU $=0.5$ is the primary model-selection metric.

\paragraph{VisDrone-overlap closed-set stress test}
The broader business corpus uses a 45-category company ontology. Within the
delivered Synthetic corpus, eight categories have nonzero ground-truth support
and correspond directly to VisDrone detection categories:
\texttt{pedestrian}, \texttt{bicycle}, \texttt{car}, \texttt{van},
\texttt{truck}, \texttt{tricycle}, \texttt{awning-tricycle}, and
\texttt{motor}. The company ontology also contains \texttt{people} and
\texttt{bus}, but no instances of these two categories are annotated in the
delivered Synthetic corpus. We therefore evaluate the eight empirically
supported classes while preserving their canonical VisDrone category
identifiers rather than renumbering them into a new contiguous label space.

The Real reference uses the complete 1,610-image VisDrone2019-DET test-dev
split~\cite{zhu2022visdrone}. The Synthetic reference is the same 446-image
audited business-scenario corpus used in the distribution analysis, comprising
80 fire/smoke, 87 illegal-fishing, 100 illegal-parking, 91 water-rescue, and
88 street-vending images. All 446 Synthetic images are retained in evaluation:
340 contain at least one aligned traffic-class ground-truth instance, while
106 contain no GT annotations in the retained eight-class evaluation slice. After ontology alignment, the
Synthetic set contains 6,478 ground-truth boxes over the eight supported
classes.

Company-source labels are mapped into the same zero-based canonical VisDrone
label space used by the Real annotations and model predictions:
\texttt{pedestrian} $9\!\rightarrow\!0$,
\texttt{bicycle} $11\!\rightarrow\!2$,
\texttt{car} $12\!\rightarrow\!3$,
\texttt{van} $13\!\rightarrow\!4$,
\texttt{truck} $14\!\rightarrow\!5$,
\texttt{tricycle} $15\!\rightarrow\!6$,
\texttt{awning-tricycle} $16\!\rightarrow\!7$, and
\texttt{motor} $18\!\rightarrow\!9$.
Real annotations and detector outputs are converted from their respective
native label conventions to the same canonical IDs. Scoring is restricted
to the eight retained categories on both test sets.

We evaluate four pretrained closed-set detectors:
YOLOv8m, YOLO11m, RF-DETR Medium~\cite{robinson2026rfdetr},
and Faster R-CNN R50-FPN~\cite{ren2015fasterrcnn}.
YOLOv8m and YOLO11m use public VisDrone model-zoo checkpoints,
RF-DETR Medium uses the corresponding VisDrone release with the
DetectionBench evaluation interface, and Faster R-CNN uses the VisDrone
checkpoint released with VLM4RSDet~\cite{shi2026vlm4rsdet}
(hereafter Faster R-CNN).
\footnote{Repositories: VisDrone Toolkit,
\url{https://github.com/dronefreak/VisDrone-dataset-python-toolkit};
DetectionBench, \url{https://github.com/dronefreak/DetectionBench};
VLM4RSDet, \url{https://github.com/cszzshi/VLM4RSDet}.}
All checkpoints are external to the Synthetic benchmark and are used in
inference-only mode.

For every detector, preprocessing, input resolution, confidence and
post-processing policy, class mapping, and evaluator are held fixed between
Real and Synthetic evaluation. Predictions are converted to the same canonical
COCO representation and evaluated with a common \texttt{pycocotools}-based
protocol. In the native VisDrone annotations, category-0 regions are retained
as ignore regions and score-zero ground-truth objects are ignored. Ground-truth
and predicted boxes with at least 50\% of their area inside an ignore region
are removed before COCO evaluation.

COCO AP over IoU thresholds $0.50{:}0.05{:}0.95$ is the primary
model-selection metric. Evaluation uses
\texttt{maxDets=[1,10,300]}, so the reported AP uses at most 300 detections
per image; AP50, AP75, and AR@300 are retained as secondary diagnostics.

\begin{figure*}[!t]
    \centering
    \includegraphics[width=0.92\textwidth]
    {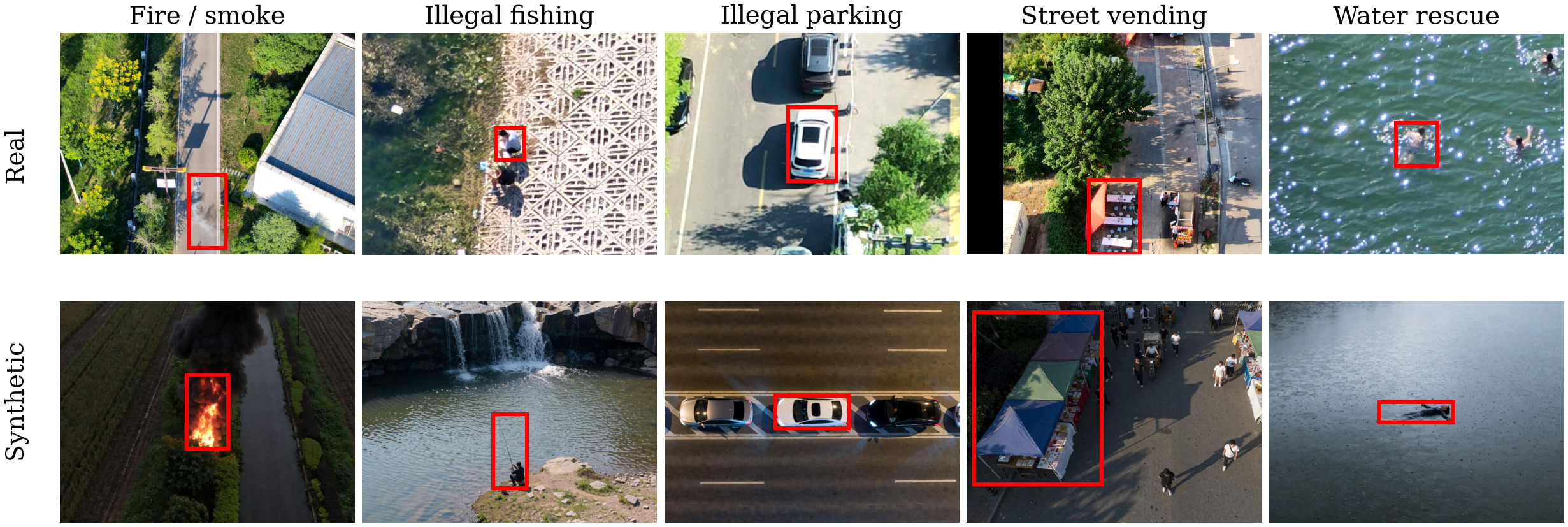}
    \caption{
    Representative Real and Synthetic examples from the five business-scenario
    families. The examples illustrate semantic scenario coverage in the
    separately materialized business corpus but do not establish
    Real--Synthetic evaluation validity.
    }
    \label{fig:scenario-breadth}
\end{figure*}

\paragraph{Cross-scenario audit corpus}
The distribution audit uses the same five-family, 446-image Synthetic corpus
together with 194 Real business-scenario images. The combined 45-category
audit contains 8,867 annotated objects and is kept separate from the
eight-class detector evaluation. Figure~\ref{fig:scenario-breadth} shows representative Real and Synthetic
examples from the five business families.

\subsection{Requirement-Specific Success Case: Road-Debris Validation}
\label{sec:road-debris-validation}

\paragraph{Performance and point-estimate agreement}
Table~\ref{tab:road-debris-main} reports the common road-debris metrics.
All four systems obtain lower F1 on Synthetic data, with decreases ranging
from $-0.076$ for Gemini to $-0.205$ for Rex-Omni. Thus, Synthetic scores are
not numerically interchangeable with Real scores.

\begin{table}[!htbp]
\centering
\caption{Road-debris performance at IoU $=0.5$ (Real/Synthetic).}
\label{tab:road-debris-main}
\scriptsize
\setlength{\tabcolsep}{2.2pt}
\renewcommand{\arraystretch}{1.05}
\begin{tabular}{lcccc}
\toprule
Model & P & R & F1 & $\Delta F_1$ \\
\midrule
Gemini
& 0.928/0.734
& 0.619/0.610
& \textbf{0.742/0.667}
& -0.076 \\

Qwen3-VL
& 0.628/0.466
& 0.477/0.421
& 0.542/0.443
& -0.100 \\

LocateAnything
& 0.843/0.440
& 0.374/0.358
& 0.518/0.395
& -0.124 \\

Rex-Omni
& 0.541/0.188
& 0.410/0.431
& 0.467/0.262
& -0.205 \\
\bottomrule
\end{tabular}
\end{table}

\begin{figure}[!htbp]
    \centering
    \includegraphics[width=\columnwidth]
    {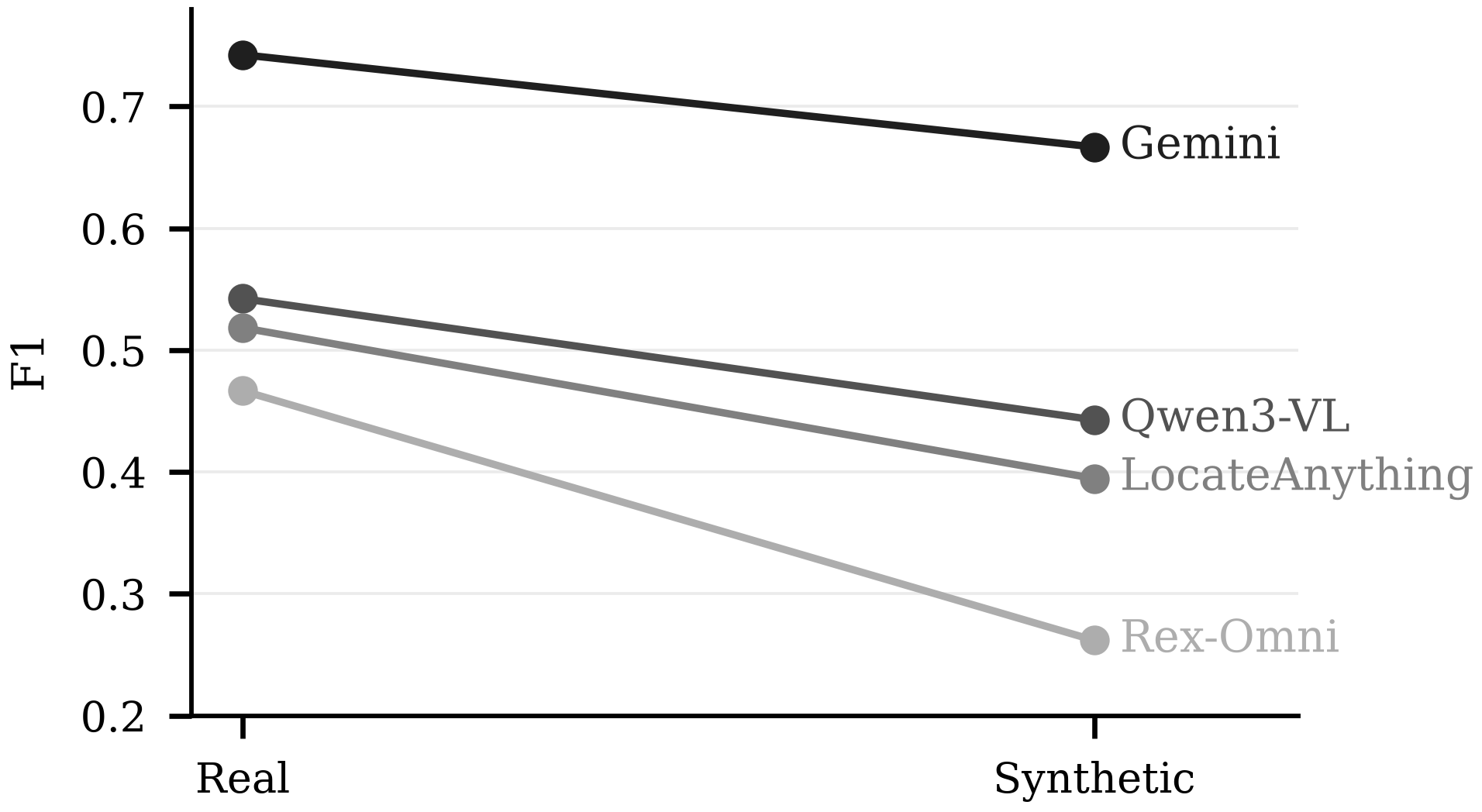}
    \caption{
Model-wise Real--Synthetic micro-F1 comparison for the four road-debris
SUTs. Dots show point estimates; all Synthetic scores are lower while the
ordering remains unchanged.
}
    \label{fig:real-syn-f1}
\end{figure}

Despite these absolute shifts, the point-estimate ordering is identical:
Gemini $>$ Qwen3-VL $>$ LocateAnything $>$ Rex-Omni.
Figure~\ref{fig:real-syn-f1} shows that all four point estimates decrease on
Synthetic data, while the ordering remains unchanged.

\paragraph{Uncertainty and size sensitivity}
Among the six pairwise comparisons, bootstrap analysis classifies three as
preserved, none as reversed, and three as unresolved under the 95\% intervals.
The resolved comparisons are Gemini versus each of the other three systems.
Gemini is top-ranked in all 10,000 bootstrap replicates on both Real and
Synthetic data.

The top-model choice is also unchanged in the size-matched sensitivity
analysis. When the 200-image Synthetic set
is repeatedly subsampled without replacement to 76 images, Gemini remains
top-ranked in all 10,000 repetitions. The complete four-model ordering matches
the Real point estimate in 71.44\% of subsets, showing that top-model selection
is substantially more stable than the lower-rank ordering.

\begin{figure*}[!t]
    \centering
    \includegraphics[width=0.88\textwidth]
    {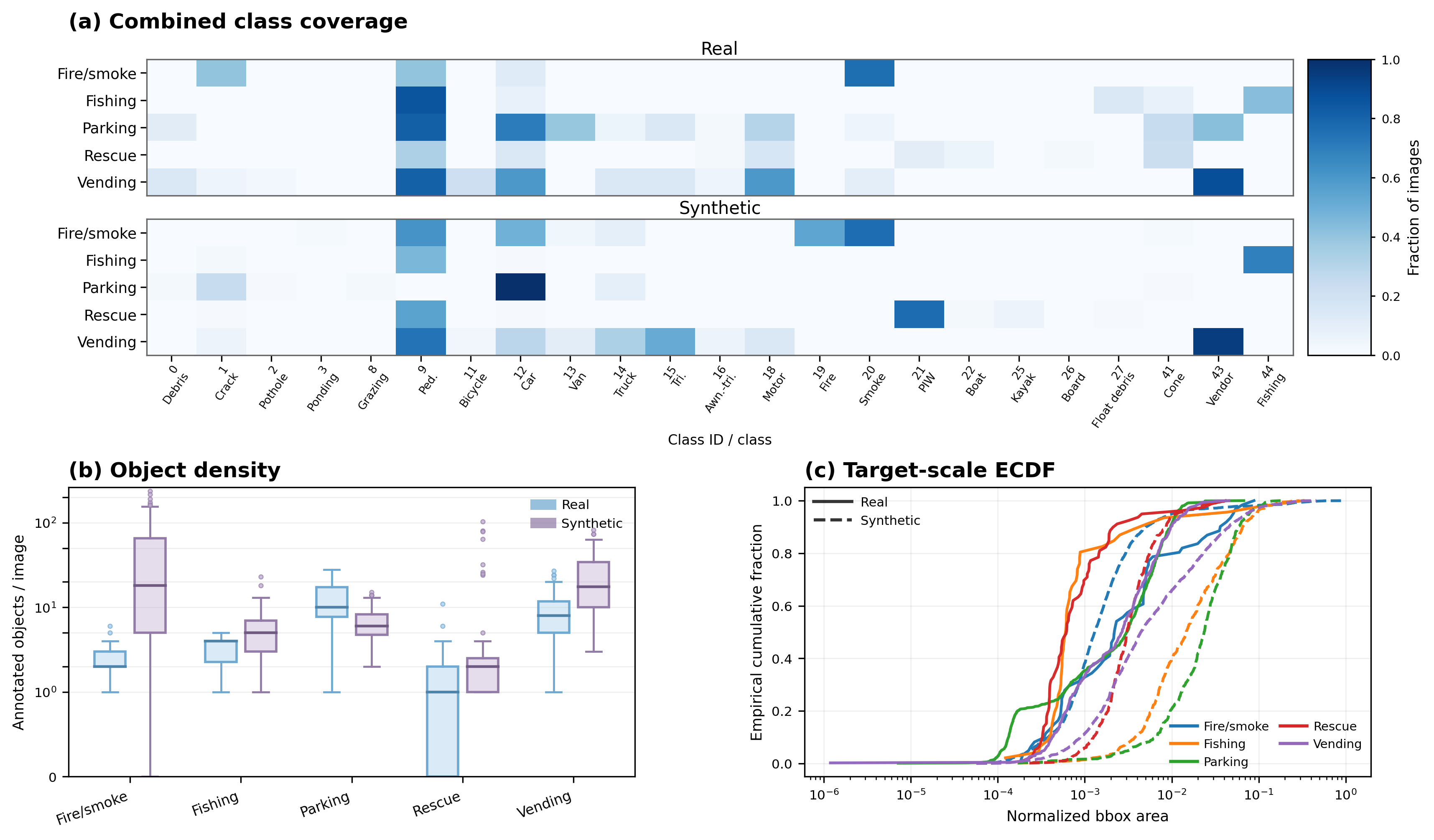}
    \caption{
Real--Synthetic distribution audit across five UAV business families:
(a) image-level class prevalence, (b) annotated-object density, and
(c) ECDF of normalized bounding-box area.
}
    \label{fig:distribution-audit}
\end{figure*}

\paragraph{Model-dependent failure behavior}
Preserved model selection does not imply identical error behavior.
Table~\ref{tab:failure-analysis} decomposes the shift into matched detections,
false positives, empty outputs, and localization quality.
Zero-prediction rate denotes images with no retained predictions after
parsing and validity checks; unparseable responses therefore count as
zero-prediction images.

\begin{table}[!htbp]
\centering
\caption{
Road-debris diagnostics (Real/Synthetic).
TP and FP are per image; IoU is mean IoU over matched TPs.
}
\label{tab:failure-analysis}

\scriptsize
\setlength{\tabcolsep}{1.5pt}
\renewcommand{\arraystretch}{1.05}

\begin{tabular}{@{}lcccc@{}}
\toprule
Model & TP/img & FP/img & Zero \% & Matched IoU \\
\midrule
Gemini
& 3.39/3.36
& 0.26/1.22
& 1.3/1.0
& 0.814/0.816 \\

Qwen3-VL
& 2.62/2.32
& 1.55/2.66
& 7.9/7.0
& 0.771/0.782 \\

LocateAnything
& 2.05/1.97
& 0.38/2.51
& 28.9/35.0
& 0.792/0.773 \\

Rex-Omni
& 2.25/2.37
& 1.91/10.24
& 2.6/0.0
& 0.745/0.755 \\
\bottomrule
\end{tabular}
\end{table}

Synthetic evaluation increases false positives for all four SUTs, whereas
matched-box IoU remains comparatively stable. The strongest change occurs for
Rex-Omni, whose FP/image increases from 1.91 to 10.24 while TP/image remains
similar. The road-debris benchmark therefore preserves the top-model decision
without reproducing either the same absolute scores or identical failure
profiles.

\subsection{VisDrone-Overlap Stress Test}
\label{sec:visdrone-overlap}

The separately materialized business corpus described in
Sec.~\ref{sec:experimental-setup} provides a second model-selection case.
Its eight VisDrone-compatible categories with nonzero Synthetic GT support
define the closed-set evaluation slice.

\paragraph{Closed-set model-selection disagreement}
Using the four closed-set detectors defined in
Sec.~\ref{sec:experimental-setup}, we apply the fixed validation protocol of
Sec.~\ref{sec:protocol}. Table~\ref{tab:visdrone-overlap} reports the Real and
Synthetic results. All four detectors achieve higher AP on the Synthetic
benchmark, but the gains are strongly model-dependent.
\begin{table}[!htbp]
\centering
\caption{VisDrone-overlap performance (Real/Synthetic, 0--100).
Rank is by AP; \texttt{maxDets}=300.}
\label{tab:visdrone-overlap}
\scriptsize
\setlength{\tabcolsep}{1.8pt}
\renewcommand{\arraystretch}{1.05}
\begin{tabular}{lccccc}
\toprule
Model & AP & AP50 & AP75 & AR@300 & Rank \\
\midrule
Faster R-CNN
& 20.45/26.37
& 38.88/39.43
& 19.34/26.92
& 36.71/51.43
& 1/4 \\

RF-DETR
& 19.13/39.12
& 35.94/51.78
& 17.92/41.90
& 33.96/71.55
& 2/1 \\

YOLO11m
& 18.99/31.37
& 33.18/41.01
& 19.21/33.59
& 29.51/46.78
& 3/2 \\

YOLOv8m
& 18.25/31.23
& 31.63/40.18
& 18.45/32.33
& 28.22/48.40
& 4/3 \\
\bottomrule
\end{tabular}
\end{table}

More importantly, the model-selection conclusion changes. The Real reference
ranks Faster R-CNN first, followed by RF-DETR, YOLO11m, and YOLOv8m, whereas
Synthetic evaluation ranks RF-DETR first and Faster R-CNN last. The top-ranked
system changes, and three of the six pairwise point-estimate preference
directions reverse, all involving Faster R-CNN.

The point-estimate disagreement shows that shared label support alone does not
guarantee the same model-selection outcome. We next characterize the same
Synthetic business corpus through a separate distributional audit.

\subsection{Cross-Scenario Distribution Audit}
\label{sec:distribution-results}

This audit uses the same 446-image Synthetic business-scenario corpus as the
VisDrone-overlap stress test, but compares it with the corresponding Real
business-scenario collections rather than with VisDrone test-dev. It therefore
characterizes distributional discrepancies in the constructed business corpus
and is not a causal analysis of the detector-ranking disagreement.
We compare annotated class prevalence, object density, and target scale
across the five business-scenario families. Figure~\ref{fig:distribution-audit}
summarizes the principal differences.

\paragraph{Composition and density}
All five business families are represented on both sides, but their internal
composition differs substantially. For example, pedestrian annotations are present in 81.25\% of Real illegal-parking images,
whereas no pedestrian annotations are present in the corresponding Synthetic
subset; car annotations are present in 70.31\% and 100\%, respectively.
Object-density shifts are also scenario-dependent: Synthetic fire/smoke and
street-vending scenes are denser, whereas Real illegal-parking scenes contain
more annotated objects.

\paragraph{Target scale}
Using normalized bounding-box area,
$a_{\mathrm{norm}}=(w/W)(h/H)$, and each Real family's $Q_{25}$ area as a
small-target reference, all five Synthetic families under-represent the
smallest-target regime. The difference is especially large for illegal
fishing, illegal parking, and water rescue, where only 0.78\%, 1.05\%, and
0.35\% of Synthetic boxes fall within the corresponding Real-defined
small-target regime.

\paragraph{Observed distributional gaps}
The audit shows that semantically appropriate Synthetic scenes can differ
substantially from the corresponding Real business-scenario collections in
annotated class prevalence, density, and target scale. These observations
document coverage gaps in the constructed corpus rather than a causal
explanation for the detector-ranking disagreement in
Sec.~\ref{sec:visdrone-overlap}.

\section{Discussion and Limitations}
\label{sec:discussion}

\paragraph{Synthetic benchmark trust is decision-specific}
The two cases show that benchmark trust is decision-specific. The road-debris
Synthetic benchmark preserves the top-model decision despite systematic
absolute-score shifts and uncertainty in the lower-rank ordering. In contrast,
the VisDrone-overlap stress test changes the point-estimate preferred detector
despite evaluation over the same eight supported VisDrone-compatible classes. Thus, neither numerical
score similarity nor semantic label alignment alone establishes Synthetic
benchmark validity.

\paragraph{Observed distributional gaps in the constructed benchmarks}
The cross-scenario audit reveals substantial Real--Synthetic
differences in class composition, density, and target scale, with
all five Synthetic families under-representing the smallest-target regime.
Semantic realization of the requested event therefore does not guarantee
faithful reproduction of the perception conditions under which that event
occurs in Real UAV data. We interpret these mismatches diagnostically rather
than causally: the present experiments do not establish that any individual
factor causes a particular model-preference change.

\paragraph{Limitations and intended use}
The two cases differ in task, Real reference, SUT panel, Synthetic corpus,
and generation backend, and therefore are complementary case studies rather
than a controlled ablation of generation fidelity. The
road-debris Real set is modest and independently sampled from the Synthetic
set, although bootstrap and size-matched analyses quantify finite-sample
uncertainty. The VisDrone-overlap case establishes point-estimate model-selection
disagreement but does not provide the uncertainty-aware pairwise
classification reported for road debris. In
addition, the generation backend is treated as a black box, and we evaluate
perception components rather than complete UAV autonomy or operational safety.

Synthetic UAV test sets should therefore complement rather than replace Real
evaluation. Their validity must be established for the specific task,
operating regime, ontology, and model panel of interest.
\section{Conclusion}
\label{sec:conclusion}

We studied whether Synthetic UAV test sets preserve Real-data model-selection
conclusions. The road-debris case preserves the top-model choice under
bootstrap and size-matched analysis, whereas the VisDrone-overlap case changes
the point-estimate preferred detector. A separate five-scenario audit documents
Real--Synthetic differences in class prevalence, density, and target scale.
These results support a decision-specific view of benchmark trust: semantic
plausibility or label overlap is insufficient, and validity must be established
empirically for the target task and model panel.

\section*{Acknowledgment}
\textit{Generative AI use:}
Synthetic imagery for the road-debris dataset was generated
using Nano Banana Pro; the business corpus was generated
using a separate internal backend.
All technical claims and experimental results were reviewed
by the authors.

\end{document}